\documentclass[twocolumn,10pt]{asme2ej}

\input{paper_style.sty}
\usepackage{graphicx} %% for loading jpg figures
\usepackage{hyperref}   % to set up hyperlinks
\hypersetup{
	colorlinks=true,
	linkcolor=blue,
	citecolor=blue,
	urlcolor=blue,
}
\usepackage[square,numbers]{natbib}

\title{Displacement-Actuated Continuum Robots:\\A Joint Space Abstraction}

\author{Reinhard M.~Grassmann
    \affiliation{
	Continuum Robotics Laboratory\\
	Department of Mathematical \& \\Computational Sciences\\
	University of Toronto\\
	Mississauga, ON L5L 1C6, Canada\\
    Email: reinhard.grassmann@utoronto.ca
    }
}

\author{Jessica Burgner-Kahrs
    \affiliation{
	Continuum Robotics Laboratory\\
	Department of Mathematical \& \\Computational Sciences\\
	University of Toronto\\
	Mississauga, ON L5L 1C6, Canada\\
    Email: jbk@cs.toronto.edu
    }
}

\begin{document}

\maketitle    

%%%%%%%%%%%%%%%%%%%%%%%%%%%%%%%%%%%%%%%%%%%%%%%%%%%%%%%%%%%%%%%%%%%%%%
\begin{abstract}
{\it 
The displacement-actuated continuum robot as an abstraction has been shown as a key abstraction to significantly simplify and improve approaches due to its relation to the Clarke transform.
To highlight further potentials, we revisit and extend this abstraction that features an increasingly popular length extension and an underutilized twisting.
For each extension, the corresponding mapping from the joint values to the local coordinates of the manifold embedded in the joint spaces is provided. 
Each mapping is characterized by its compactness and linearity.
}
\end{abstract}

%%%%%%%%%%%%%%%%%%%%%%%%%%%%%%%%%%%%%%%%%%%%%%%%%%%%%%%%%%%%%%%%%%%%%%
% \begin{nomenclature}
% \entry{A}{You may include nomenclature here.}
% \entry{$\alpha$}{There are two arguments for each entry of the nomemclature environment, the symbol and the definition.}
% \end{nomenclature}

% The primary text heading is  boldface and flushed left with the left margin.  The spacing between the  text and the heading is two line spaces.

%%%%%%%%%%%%%%%%%%%%%%%%%%%%%%%%%%%%%%%%%%%%%%%%%%%%%%%%%%%%%%%%%%%%%%

% Introduction
\section{Introduction}

Utilizing a suitable abstraction is arguably a stepping stone to breaking through unsolved problems, developing advanced methods, and gaining further insights.
For kinematic modeling in robotics, the pose is often decomposed in a translational vector and an orientation representation, where different representation exists, \textit{e.g.}, as a rotation matrix, one of twelve sets of Euler angles, or quaternions.
This formulation of the pose is an abstraction of the task space.
Coupled abstraction exists, too, \textit{e.g.}, dual quaternions.
Especially in serial-kinematic rigid robots, an intermediate space based on line geometry is used to bridge the task space and joint space, such as the Denavit-Hartenberg parameters or the product of exponentials.
Due to the nature of serial-kinematic rigid robots, the joint space is the configuration space, and no special abstraction is needed.
Furthermore, the joint and actuation space are normally linearly related through a gear ratio.
However, for serial-kinematic robots with flexible joints, \textit{e.g.}, \cite{Albu-SchaefferHirzinger_et_al_IRIJ_2007, HaddadinHaddadin_et_al_RAM_2022}, the joint flexibly has to be considered, \textit{e.g.}, by extending the states using a spring model \cite{DeLucaBook_HOR_2016} leading to a $4\textsuperscript{th}$ order dynamic flexible joint robot.

\begin{figure*}
    \centering
    \includegraphics[width=\textwidth]{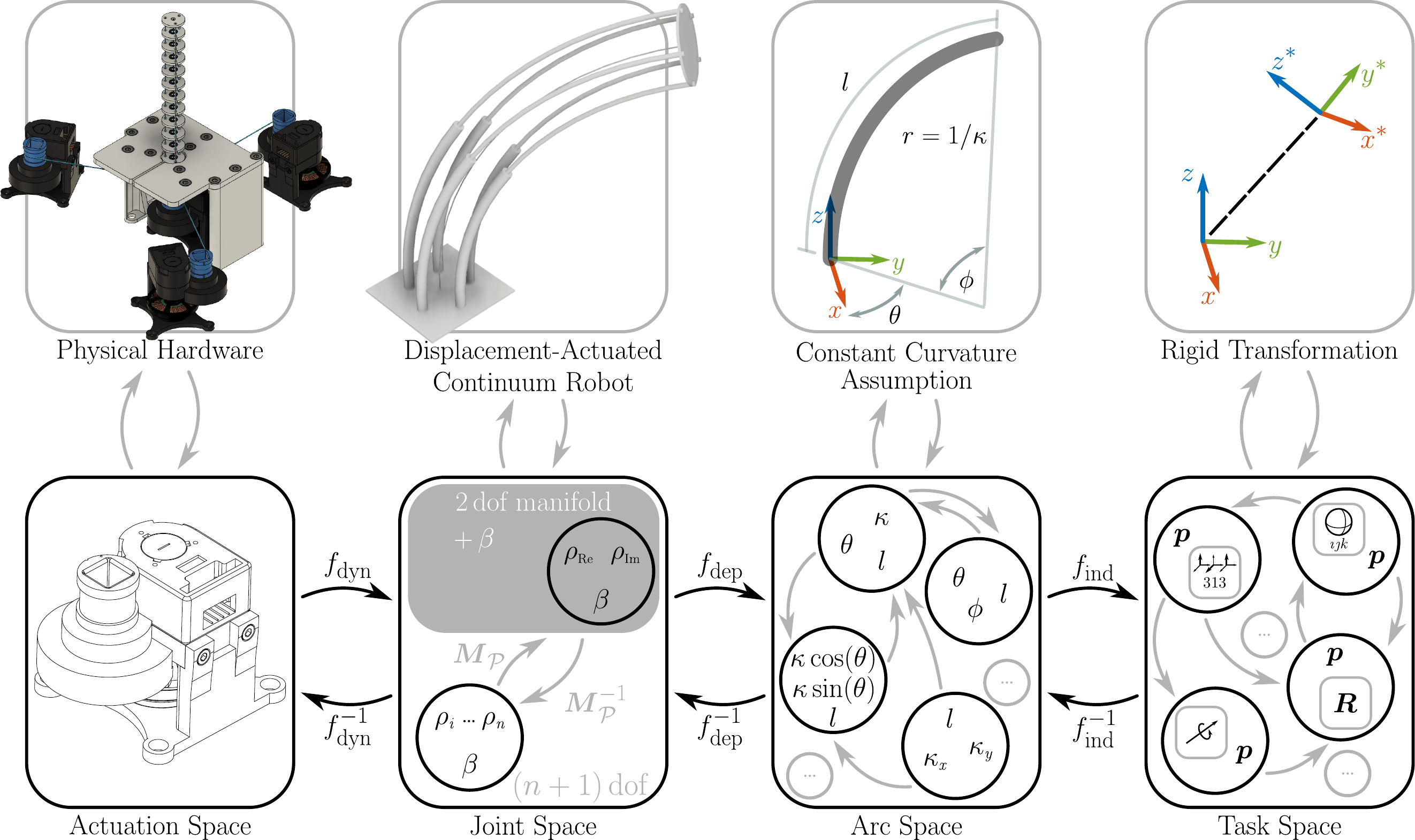}
    \vspace{-1em}
    \caption{
        Abstraction leads to mathematical models.
        The model of the physical world informs the abstraction and this representation.
    }
    \label{fig:abstraction_overview}
\end{figure*}

A similar approach is used for kinematic modeling in continuum and soft robotics, see Fig.~\ref{fig:abstraction_overview}.
For obvious reasons, the abstraction of the task space is similar. 
Notice that one can observe that most approaches are limited to position, highlighting the challenges in considering the properties of $SO(3)$.
The intermediate space often utilizes geometry based on circles and arcs. 
Hence, one might prefer the arc space \cite{MarcheseKatzschmannRus_IROS_2014, GrassmannBurgner-Kahrs_et_al_Frontiers_2022} over the configuration space \cite{WebsterJones_IJRR_2010, RaoBurgner-Kahrs_et_al_Frontiers_2021} to denote this intermediate space.
In contrast to serial-kinematic rigid robots, the joints are often interdependent and, at the same time, the dimensionality of the joint space is higher than the accessible degrees of freedom in the arc space as well as the task space.
Both characteristics impose challenges in modeling.
To overcome this, improved state representations \cite{DellaSantinaBicchiRus_RAL_2020, AllenAlbert_et_al_RoboSoft_2020, DianGuo_et_al_Access_2022} have been introduced.
However, these are limited to three or four joints and their symmetrical location.
This hints at a missing abstraction, as the improved state representations cannot be extended to overcome their limitations.
The actuation space is often used as a synonym for the joint space.
Although the distinction might not be necessary, since most continuum robot prototypes use stiff actuators, \textit{e.g.}, DC motors with high gear ratio, it can be important to do so when considering quasi-drive actuation, \textit{e.g.}, \cite{GrassmannBurgner-Kahrs_et_al_Frontiers_2024}, that is back-drivable and capable of proprioceptive sensing.

From the above comparison, we conclude that an abstraction for the joint space is missing.
This gap causes restrictions on design consideration, creates confusion on the system's redundancy, favors brute forcing a solution, constrains approaches to arc space, and obstructs insights, limiting the research endeavor and applicability of continuum robots.
Furthermore, it causes the actuation space and joint space to be improperly abstracted.
We note that if the joint space is properly abstracted, the actuation space is automatically abstracted as well.
The abstraction of the actuation space depends on the actuation used, \textit{e.g.}, pneumatic bellows, cable, tendons, and push-pull rods.

Surprisingly, a rather straightforward abstraction can address most of the challenges.
This abstraction was recently introduced by Grassmann \textit{et al.} \cite{GrassmannSenykBurgner-Kahrs_arXiv_2024} to generalize the improved state representations.
The so-called \textit{displacement-actuated continuum robot} is a necessary abstraction to achieve the generalization.
While this abstraction has been defined for symmetrical joint locations, the same group has extended their approach \cite{GrassmannBurgner-Kahrs_arXiv_2024a} to account for random joints.
However, the current abstraction only accounts for $\SI{2}{dof}$ per segment, although $\SI{4}{dof}$ per segment can be achieved \cite{GrassmannBurgner-Kahrs_et_al_Frontiers_2022}. 

In a recent work by Grassmann \& Burgner-Kahrs \cite{GrassmannBurgner-Kahrs_arXiv_2025}, they show that Clarke coordinates \cite{GrassmannSenykBurgner-Kahrs_arXiv_2024} are linear to improved state parameterizations proposed by Della \textit{et al.} \cite{DellaSantinaBicchiRus_RAL_2020}, Allen \textit{et al.} \cite{AllenAlbert_et_al_RoboSoft_2020}, and Dian \textit{et al.} \cite{DianGuo_et_al_Access_2022}.
Therefore, methods based on approaches such as forward kinematics \cite{DellaSantinaBicchiRus_RAL_2020, DianGuo_et_al_Access_2022, GrassmannSenykBurgner-Kahrs_arXiv_2024}, inverse kinematics \cite{GrassmannSenykBurgner-Kahrs_arXiv_2024}, Jacobian of the forward kinematics \cite{DellaSantinaBicchiRus_RAL_2020, DianGuo_et_al_Access_2022}, Lagrangian dynamics \cite{DellaSantinaBicchiRus_RAL_2020, DianGuo_et_al_Access_2022}, kinematic control \cite{GrassmannSenykBurgner-Kahrs_arXiv_2024, GrassmannBurgner-Kahrs_arXiv_2024a}, model-based control \cite{DellaSantinaBicchiRus_RAL_2020}, sliding mode control \cite{DianGuo_et_al_Access_2022}, state observer \cite{DianGuo_et_al_Access_2022}, trajectory generation \cite{GrassmannBurgner-Kahrs_arXiv_2024b}, and sampling method \cite{GrassmannSenykBurgner-Kahrs_arXiv_2024} are well-defined.
Furthermore, derived methods are linear, compact, computationally efficient, and closed-form.
While the current abstraction \cite{GrassmannSenykBurgner-Kahrs_arXiv_2024, GrassmannBurgner-Kahrs_arXiv_2024a} applies to a wide set of continuum and soft robots, it is limited to spatial bending.
The inclusion of extensible segments \cite{NguyenBurgner-Kahrs_IROS_2015, GrassmannBurgner-Kahrs_et_al_Frontiers_2022} or segments that can twist \cite{StarkeBurgner-Kahrs_et_al_IROS_2017} is desirable.
The former has become increasingly relevant \cite{AlandoliFanLiu_Robotica_2024}.

In the presented work, we are building work on a displacement-actuated continuum robot \cite{GrassmannSenykBurgner-Kahrs_arXiv_2024} to extend this abstraction to $\SI{3}{dof}$ and $\SI{4}{dof}$ per segment as well as to multi-segment design.
Furthermore, we summarize recent achievements, highlight current possibilities, and point to challenges toward a complete understanding of this abstraction.
In particular, the following contributions are made:
\begin{itemize}
    \item[$\bullet$] An in-depth description of the key abstraction \textit{Displacement-Actuated Continuum Robot}
    \item[$\bullet$] Its extension to fully actuated segment and multi-segment design
    \item[$\bullet$] Discussion on joint representation as abstraction merging from the key abstraction
\end{itemize}

% Abstractions
\section{Displacement-Actuated Continuum Robot}
\label{sec:DACR}

An abstraction helps organize the relevant details and hide the irrelevant information about the actuation, see Fig.~\ref{fig:abstraction_overview}.
The relevant details are the kinematic parameters and three assumptions.
By stating them, we gain three benefits.
First, using displacement-actuated joints for kinematics, it is irrelevant whether the displacement is caused by pneumatics, hydraulics, cable, push-pull rod, tendon, or other means.
This allows for a wide range of continuum and soft robots to be covered.
Second, it reveals implicit assumptions that might be hidden from the researcher and practitioner and might cause challenges or failures.
Third, clear questions can be formulated, \textit{e.g.}, to find a point of view to exploit the abstraction.

Grassmann \textit{et al.} \cite{GrassmannSenykBurgner-Kahrs_arXiv_2024} introduce the displacement-actuated continuum robot as an abstraction.
Here, we expand on their brief introduction, incorporate variable kinematic parameters, and introduce variants for segments with higher degrees of freedom.

\subsection{Kinematic Design Parameters}

\begin{figure}
    \centering
    \includegraphics[width=0.8\columnwidth]{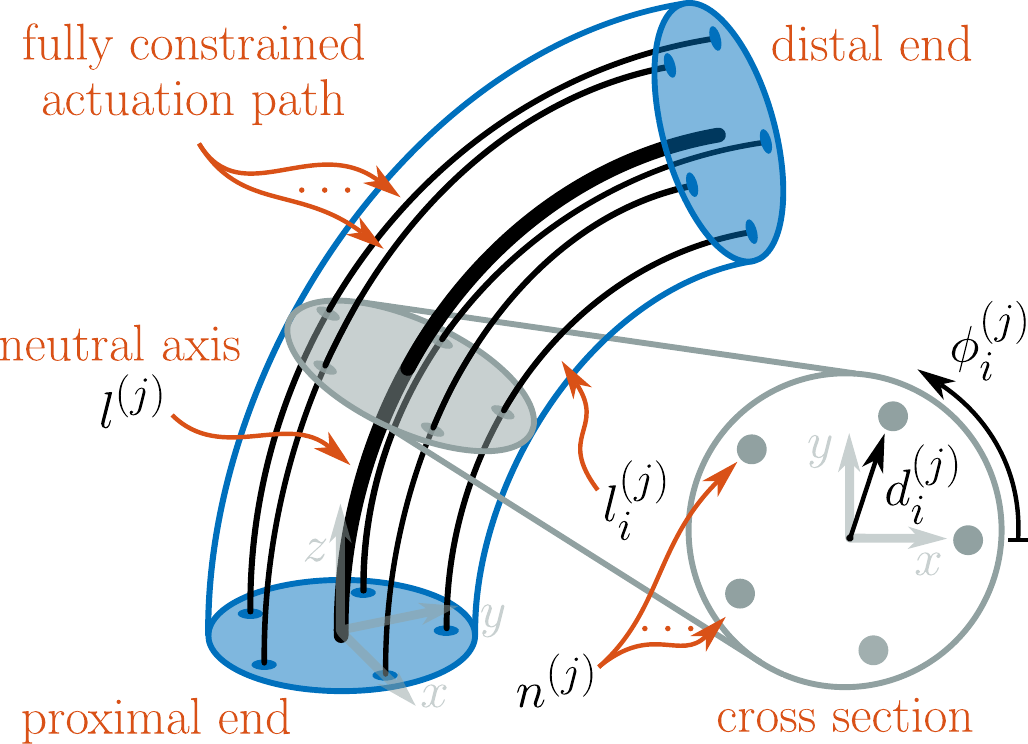}
    \caption{
        Displacement-actuated continuum robot.
        Assuming fully constrained actuation paths, the distance $d_i$ and angle $\psi_i$ of each joint location on the cross-section is constant along the arc length.
    }
    \label{fig:dacr}
\end{figure}

A set of kinematic design parameters fully describes this abstraction.
The relation between the kinematic design parameters and a DACR with one segment as a continuum structure is sketched in Fig.~\ref{fig:dacr}.
The kinematic design parameters are the number of joints $n\kthsegment{j}$, the initial length $l\kthsegment{j}$, the distance $d\kthsegment{j}\kthjoint{i}$, and the angle $\psi\kthsegment{j}\kthjoint{i}$.
The last two parameters are the polar coordinates $\left(d\kthsegment{j}\kthjoint{i}, \psi\kthsegment{j}\kthjoint{i}\right)$ of each joint on a cross-section, see Fig.~\ref{fig:dacr}.
Furthermore, superscript $j$ and subscript $i$ indicate the relation to $j\textsuperscript{th}$ segment and $i\textsuperscript{th}$ joint, respectively.
Table~\ref{tab:kinematic_parameters} lists the kinematic design parameters with the corresponding subscript and superscript.

\begin{table}
	\centering
	\caption{
        Kinematic design parameters
    }
	\label{tab:kinematic_parameters}
	\begin{tabular}{@{} r  p{6.5cm} @{}}
		\toprule
		\multicolumn{1}{N}{Notation}
		& \multicolumn{1}{N}{Description}\\
		\cmidrule(r){1-1}
		\cmidrule(l){2-2}
	    $l\kthsegment{j}$ & Length of the $j\textsuperscript{th}$ segment\\[.5em]
	    $n\kthsegment{j}$ & Number of joints of the $j\textsuperscript{th}$ segment\\[.5em]
	    $\psi\kthsegment{j}\kthjoint{i}$ & angular part of the polar coordinate $\left(d\kthsegment{j}\kthjoint{i}, \psi\kthsegment{j}\kthjoint{i}\right)$ of the $i\textsuperscript{th}$ joint location of the $j\textsuperscript{th}$ segment\\[.5em]
	    $d\kthsegment{j}\kthjoint{i}$ & translational part of the polar coordinate $\left(d\kthsegment{j}\kthjoint{i}, \psi\kthsegment{j}\kthjoint{i}\right)$ of the $i\textsuperscript{th}$ joint location of the $j\textsuperscript{th}$ segment\\
		\bottomrule
	\end{tabular}
\end{table}

\subsection{Segment Types and Joint Parameters}

Aside from the kinematic design parameters, joint parameters change the continuum manipulator from an initial configuration to a desired one.
To introduce variants of DACR, we adapt the idea of a fully actuated segment \cite{GrassmannBurgner-Kahrs_et_al_Frontiers_2022} for DACR.
For tendon-driven continuum robots, Grassmann \textit{et al.} \cite{GrassmannBurgner-Kahrs_et_al_Frontiers_2022} introduce the concept of floating spacer disks, where the degrees of freedom of each disk classify its type.
The type of spacer disk further gives rise to the degrees of freedom of the segment.
We adopt this line of thought.
Hence, for a spatial segment of a DACR, we define:
\begin{itemize}
    \item[$\bullet$] Type-0 segment is an incompressible and torsional stiff segment that can bend in two orthogonal bending planes.
    \item[$\bullet$] Type-I segment is a torsional stiff segment that can bend and change its length.
    \item[$\bullet$] Type-II segment is an incompressible segment that can bend and twist.
    \item[$\bullet$] Type-III segment is a segment that can bend, twist, and change its length.
\end{itemize}

Note that an incompressible segment theoretically and mathematically cannot bend.
However, we assume that some compression and decompression are possible such that bending is achievable and the length of the neutral axis $l\kthsegment{j}$ is unchanged.
Furthermore, note that a type-III segment can be seen as a generalized fully actuated segment proposed by Grassmann \textit{et al.} \cite{GrassmannBurgner-Kahrs_et_al_Frontiers_2022}, where the generalization lies in changing tendon actuation to a displacement-actuated joint.
Finally, note that our notation does not account for a non-bending segment that can twist and change its length.
One might refer to it as a non-bending type-III segment.

Each of the segments can bend induced by displacements $\rhovec\kthsegment{j}$.
A segment that can change its length has an additional joint denoted by $\beta\kthsegment{j}$.
For a segment that can twist around on its neutral axis, a rotational angle $\alpha\kthsegment{j}$ at its proximal end is the associated joint.
As for the kinematic design parameters, subscript $j$ and superscript $i$ refer to $j\textsuperscript{th}$ segment and $i\textsuperscript{th}$ joint, respectively.
Table~\ref{tab:joint_parameters} lists the joint parameters.

\begin{table}
	\centering
	\caption{
        Joint parameters
    }
	\label{tab:joint_parameters}
	\begin{tabular}{@{} r  p{6.5cm} @{}}
		\toprule
		\multicolumn{1}{N}{Notation}
		& \multicolumn{1}{N}{Description}\\
		\cmidrule(r){1-1}
		\cmidrule(l){2-2}
	    $\rhovec\kthsegment{j}$ & displacements of the $j\textsuperscript{th}$ segment as vector\\[.5em]
	    $\rho\kthsegment{j}\kthjoint{i}$ & $i\textsuperscript{th}$ displacements of $\rhovec\kthsegment{j}$ as a scaler\\[.5em]
	    $\beta\kthsegment{j}$ & translational parameter of the $j\textsuperscript{th}$ segment as a scaler\\[.5em]
	    $\alpha\kthsegment{j}$ & rotational angle of the $j\textsuperscript{th}$ segment at its proximal end as a scaler\\
		\bottomrule
	\end{tabular}
\end{table}

\subsection{Assumptions}

This abstraction comes with two assumptions: a smooth backbone and a fully constrained actuation path.
The latter is an adopted terminology by Rao \textit{et al}. \cite{RaoBurgner-Kahrs_et_al_Frontiers_2021}.
Notice that, by providing a relation between a set of kinematic parameters, this can be seen as an assumption given as an equation.

\begin{figure}
    \centering
    \includegraphics[width=0.8\columnwidth]{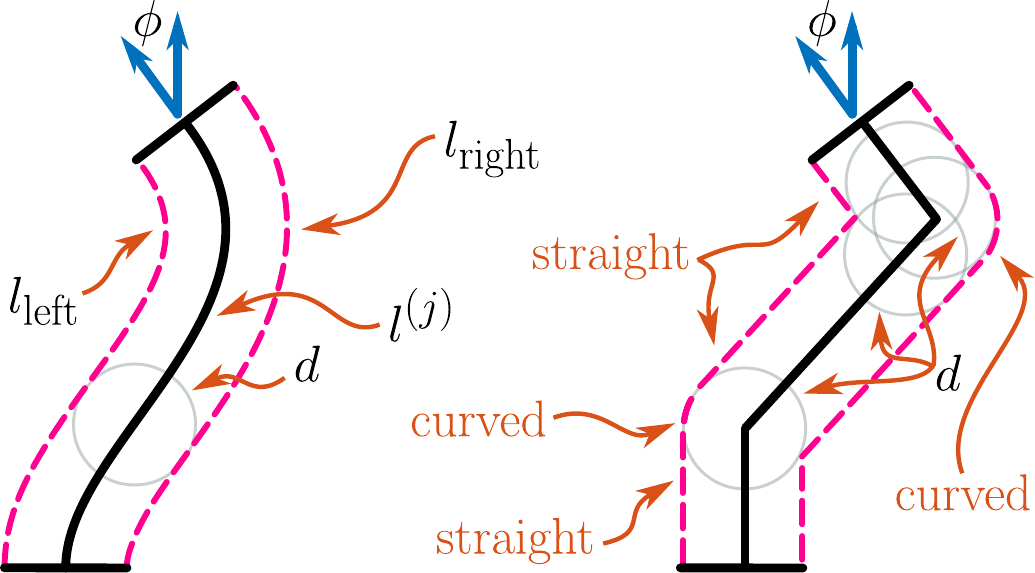}
    \caption{
        Geometric smoothness $\mathcal{G}$.
        The smooth backbone informs both fully constrained actuation paths that are parallel curves.
        The minimal distance between them is always $d$.
        Left side: The backbone of length $l\kthsegment{j}$ is a curve with continuous tangent vectors.
        Here, $l_\text{left} < l\kthsegment{j}$ and $l_\text{right} > l\kthsegment{j}$ differs to $l\kthsegment{j}$ by a same absolute among, \textit{i.e.}, $|d\phi|$.
        Right side: The construction of the parallel curves creates degenerate cases with loops and cusps between straight lines, see Pham \cite{Pham_CAD_1992} for visualizations.
        Furthermore, the fully constrained actuation paths are composed of straight and curved pieces.
        }
    \label{fig:backbone_smoothness_analytic}
\end{figure}

A smooth backbone is a crucial assumption otherwise, the robot is not considered a continuum robot according to Burgner-Kahrs \textit{et al.} \cite{Burgner-KahrsRuckerChoset_TRO_2015}.
A backbone with a continuous tangent vector is at least $\mathcal{G}^{1}$ smooth.
In contrast, a piece-wise linear backbone with at least two pieces in a non-straight configuration is $\mathcal{G}^{0}$ smooth and, therefore, it is geometrically not smooth enough.
Figure~\ref{fig:backbone_smoothness_analytic} illustrates both geometrical smooth backbones.
Note that a combination of linear and circular pieces can never be $\mathcal{G}^{2}$ smooth and, in general, it is not $\mathcal{G}^{1}$ smooth, see Biagiotti \& Melchiorri \cite{BiagiottiMelchiorri_Book_2008}, where the argument is made using analytical smoothness $\mathcal{C}$.
Fortunately, a backbone with multiple circular segments is $\mathcal{G}^{1}$ smooth if the adjoining segments have the same tangent at their intersection. 

A fully constrained actuation path, as shown in Fig.~\ref{fig:dacr} and Fig.~\ref{fig:backbone_smoothness_analytic}, is an assumption that mimics the geometric smoothness $\mathcal{G}$ of the backbone and, therefore, the nature of a continuum robot.
As illustrated in Fig.~\ref{fig:backbone_smoothness_analytic}, moving a circle or sphere with constant radius $d$ along the backbone constructs parallel curves that are fully constrained actuation paths.
While this depiction focuses on a two-dimensional case, it can be extended to the spatial case.
Let us consider the two-dimensional case depicted in Fig.~\ref{fig:backbone_smoothness_analytic} as the projection onto the bending plane, where $\theta$ is the bending plane angle, also known as the direction of the bending.
Using the idea of virtual displacement \cite{GrassmannSenykBurgner-Kahrs_arXiv_2024, FirdausVadali_AIR_2023}, the parallel curve can be interpreted as the fully constrained actuation path of the virtual displacement.
Now, we distribute this path to all $n$ joint locations using the Clarke transform \cite{GrassmannSenykBurgner-Kahrs_arXiv_2024} to complete the argument.
Note that a proper proof for the extension from Fig.~\ref{fig:backbone_smoothness_analytic} to Fig.~\ref{fig:dacr} would require the continuous formulation of the Clarke transform.

The two assumptions result in the fact that shear should not be considered and that a displacement constraint emerges. 
Due to the construction of both parallel curves, shearing of the backbone can be problematic.
Considering shear can lead to a $\mathcal{G}^0$ smooth backbone.
This justifies the four considered types, \textit{i.e.}, type-0 to type-III, at the beginning.
Moreover, the displacements being the length changes, \textit{i.e.}, $l\kthsegment{j} - l_\text{left} = \rho_\text{left}$ and $l\kthsegment{j} - l_\text{right} = \rho_\text{right}$, in Fig.~\ref{fig:backbone_smoothness_analytic} result in a constraint
\begin{align}
	\left(l\kthsegment{j} - l_\text{left}\right) + \left(l\kthsegment{j} - l_\text{right}\right) = 0, 
    \label{eq:sum_rho_planar_case}
\end{align}
which is a reformulated property of parallel curves \cite{LoriaSchuette_book_1902} that has been previously shown by Grassmann and Burgner-Kahrs \cite{GrassmannBurgner-Kahrs_arXiv_2024b}.
Therefore, assuming a fully constrained actuation path and a smooth backbone leads to a displacement constraint.
\section{DACR with one type-0 Segment}

This type is often considered in the literature.
It is fully described by the kinematic design parameters and displacements $\rhovec\kthsegment{j}$.
In the following, we describe possible considerations regarding the joint locations, \textit{i.e.}, $\phi\kthsegment{j}\kthjoint{i}$ and $d\kthsegment{j}\kthjoint{i}$.
Furthermore, the joint representation, \textit{i.e.}, displacement $\rhovec\kthsegment{j}$, can be compressed without loss of information into two variables.
The key approach is the Clarke transform, which is briefly described, too.

\subsection{Joint Locations and Joint Representation}

This variant introduced by Grassmann \textit{et al.} \cite{GrassmannSenykBurgner-Kahrs_arXiv_2024} considers $n\kthsegment{j}$ symmetric arranged joint locations given by
\begin{align}
    \psi\kthsegment{j}\kthjoint{i} = 2\pi\dfrac{i - 1}{n\kthsegment{j}}
    \quad\text{and}\quad
    d\kthsegment{j}\kthjoint{i} = d\kthsegment{j} \in \mathbb{R}^+
    \label{eq:joint_location_symmetric}
    .
\end{align}
It generalizes previous designs mostly limited to $n\kthsegment{j} \in \left[3, 4\right]$ joints.
Only a few mechanical designs consider a higher $n\kthsegment{j}$, \textit{e.g.} Olson \textit{et al.} \cite{OlsonMenguc_et_al_IJSS_2020} and Davland \textit{et al.} \cite{DalvandNahavandiHowe_TRO_2018} use $n\kthsegment{j} = 6$.
The generalization lies in the extension to $n\kthsegment{j} \geq 3$.
An extension by Grassmann \& Burgner-Kahrs \cite{GrassmannBurgner-Kahrs_arXiv_2024a} is the most generalized version of a type-0 segment.
It considers arbitrary joint locations given by
\begin{align}
    \psi\kthsegment{j}\kthjoint{i} = \left[0, 2\pi\right)
    \quad\text{and}\quad
    d\kthsegment{j}\kthjoint{i} \in \mathbb{R}^+
    \label{eq:joint_location_non-symmetric}
\end{align}
on the cross-section of a continuum structure, see Fig.~\ref{fig:dacr}.

To actuate a type-0 segment, the joint values are described by displacements $\rho\kthsegment{j}\kthjoint{i}$ and represented as
\begin{align}
    \!\!\rhovec\kthsegment{j} 
    \!=\!
    \begin{bmatrix} 
        \rho\kthsegment{j}\kthjoint{1} \\[0.75em]
        \rho\kthsegment{j}\kthjoint{2} \\[0.75em]
        \vdots \\[0.75em]
        \rho\kthsegment{j}\kthjoint{n}
    \end{bmatrix}
    \!=\!
	\begin{bmatrix}
		\rhoreal\kthsegment{j}\cos\!\left(\psi\kthsegment{j}\kthjoint{1}\right) + \rhoim\kthsegment{j}\sin\!\left(\psi\kthsegment{j}\kthjoint{1}\right) \\[0.75em]
        \rhoreal\kthsegment{j}\cos\!\left(\psi\kthsegment{j}\kthjoint{2}\right) + \rhoim\kthsegment{j}\sin\!\left(\psi\kthsegment{j}\kthjoint{2}\right) \\[0.75em]
        \vdots \\[0.75em]
        \rhoreal\kthsegment{j}\cos\!\left(\psi\kthsegment{j}\kthjoint{n}\right) + \rhoim\kthsegment{j}\sin\!\left(\psi\kthsegment{j}\kthjoint{n}\right)
	\end{bmatrix}
	\label{eq:rho}
\end{align}
with $\rhovec\kthsegment{j} \in \jointspace \subset \mathbb{R}^n$ according to Grassmann \textit{et al.} \cite{GrassmannSenykBurgner-Kahrs_arXiv_2024}.
For details on the joint space $\jointspace$, we kindly refer to \cite{GrassmannSenykBurgner-Kahrs_arXiv_2024} and \cite{GrassmannBurgner-Kahrs_arXiv_2025}.
All displacements \eqref{eq:rho} are interdependent, obey
\begin{align}
	\sum_{i=1}^{n} \rho\kthsegment{j}\kthjoint{i} = 0, 
    \label{eq:sum_rho}
    % \nonumber
\end{align}
and have only two free variables \cite{GrassmannSenykBurgner-Kahrs_arXiv_2024}.
The Clarke coordinates 
\begin{align}
    \rhoclarke\kthsegment{j} = \left[ \rhoreal\kthsegment{j}, \rhoim\kthsegment{j} \right]\transpose \in \mathbb{R}^2
    \label{eq:rho_clarke}
\end{align}
are those free variables of Eqn.~\eqref{eq:rho}.
Table~\ref{tab:clarke_coordinates} lists them.
Furthermore, Eqn.~\eqref{eq:rho_clarke} are the local coordinates of the \SI{2}{dof} manifold embedded in the $n\,\SI{}{dof}$ joint space $\jointspace$.

\begin{table}
	% \vspace*{0.5em}
	\centering
	\caption{
        Clarke coordinates of $j\textsuperscript{th}$ segment
    }
	\label{tab:clarke_coordinates}
	% \vspace*{-3mm}
	\begin{tabular}{@{} r  p{6.5cm} @{}}
		\toprule
		\multicolumn{1}{N}{Notation}
		& \multicolumn{1}{N}{Description}\\
		\cmidrule(r){1-1}
		\cmidrule(l){2-2}
	    $\rhoreal\kthsegment{j}$ & real part of the Clarke coordinates of $\rhovec\kthsegment{j}$ as a scaler\\[.5em]
	    $\rhoim\kthsegment{j}$ & imaginary part of the Clarke coordinates of $\rhovec\kthsegment{j}$ as a scaler\\
		\bottomrule
	\end{tabular}
\end{table}

\subsection{Clarke Transform}

Grassmann \textit{et al.} \cite{GrassmannSenykBurgner-Kahrs_arXiv_2024} propose Clarke transform.
To transform both representations, \textit{i.e.}, Eqn.~\eqref{eq:rho} and Eqn.~\eqref{eq:rho_clarke}, into each other, the forward and inverse Clarke transform simplify to 
\begin{align}
    \rhoclarke\kthsegment{j} &= \MP\rhovec\kthsegment{j}
    \quad\text{and}\quad
    \label{eq:forward}
    \\
    \rhovec\kthsegment{j} &= \MPinv\rhoclarke\kthsegment{j}
    \label{eq:inverse}
\end{align}
respectively.
The inverse generalized Clarke transformation matrix $\MPinv$ in Eqn.~\eqref{eq:inverse} can be found through an analogy \cite{GrassmannSenykBurgner-Kahrs_arXiv_2024} incorporating Eqn.~\eqref{eq:sum_rho}, or using vectorization \cite{GrassmannBurgner-Kahrs_arXiv_2024a} of Eqn.~\eqref{eq:rho}.
It is
\begin{align}
	\MPinv
    =
	\begin{bmatrix}
		\cos\left(\psi\kthsegment{j}\kthjoint{1}\right) & \sin\left(\psi\kthsegment{j}\kthjoint{1}\right) \\[0.75em]
        \cos\left(\psi\kthsegment{j}\kthjoint{2}\right) & \sin\left(\psi\kthsegment{j}\kthjoint{2}\right) \\[0.75em]
        \vdots & \vdots\\[0.75em]
        \cos\left(\psi\kthsegment{j}\kthjoint{n}\right) & \sin\left(\psi\kthsegment{j}\kthjoint{n}\right)
	\end{bmatrix}
    ,
	\label{eq:MP_inverse}
\end{align}
which holds for symmetrically and asymmetrically arranged joints as described by Eqn.~\eqref{eq:joint_location_symmetric} and Eqn.~\eqref{eq:joint_location_non-symmetric}, respectively.
Note that Eqn.~\eqref{eq:MP_inverse} is the right inverse of the non-square matrix \eqref{eq:MP}.
Furthermore, for the sake of readability, we omit $j$ for the notation of $\MP$ and $\MPinv$ throughout this work.

For the general case \eqref{eq:joint_location_non-symmetric} presented by Grassmann \& Burgner-Kahrs \cite{GrassmannBurgner-Kahrs_arXiv_2024a}, the generalized Clarke transformation matrix $\MP$ in Eqn.~\eqref{eq:forward} is the Moore-Penrose pseudoinverse for solving undetermined linear systems.
This leads to
\begin{align}
    \MP =& \left(\left(\MPinv\right)\transpose\MPinv\right)^{-1}\left(\MPinv\right)\transpose
    ,
    \label{eq:MP_inverse_pseudo}
\end{align}
which is an exact solution.
Note that Eqn.~\eqref{eq:inverse} is an undetermined linear system and not an overdetermined linear system.
Solving the former is an exact solution, whereas solving the latter leads to an approximation.

For symmetrical arrangement \eqref{eq:joint_location_symmetric}, Eqn.~\eqref{eq:MP_inverse_pseudo} simplifies to 
\begin{align}
    \hspace*{-0.5em} 
	\MP
    \!
    =
    \!
    \dfrac{2}{n\kthsegment{j}}
    \!\!
	\begin{bmatrix}
		\cos\!\left(\psi\kthsegment{j}\kthjoint{0}\right) &\hspace*{-0.0em} \cos\!\left(\psi\kthsegment{j}\kthjoint{1}\right) &\hspace*{-0.0em} \cdots &\hspace*{-0.0em} \cos\!\left(\psi\kthsegment{j}\kthjoint{n}\right)\\[0.75em]
		\sin\!\left(\psi\kthsegment{j}\kthjoint{0}\right) &\hspace*{-0.0em} \sin\!\left(\psi\kthsegment{j}\kthjoint{1}\right) &\hspace*{-0.0em} \cdots &\hspace*{-0.0em} \sin\!\left(\psi\kthsegment{j}\kthjoint{n}\right)
	\end{bmatrix}
    \hspace*{-0.35em} 
    % ,
	\label{eq:MP}
\end{align}
when the joint locations are symmetrically arranged according to Eqn.~\eqref{eq:joint_location_symmetric}.
Grassmann \textit{et al.} \cite{GrassmannSenykBurgner-Kahrs_arXiv_2024} propose the generalized Clarke transformation matrix \eqref{eq:MP} derived using an analogy to approaches in electrical engineering.
We kindly refer to \cite{GrassmannSenykBurgner-Kahrs_arXiv_2024, GrassmannBurgner-Kahrs_arXiv_2025} for a discussion on intuition and analogy.
Grassmann \& Burgner-Kahrs \cite{GrassmannBurgner-Kahrs_arXiv_2024a} prove the simplification from Eqn.~\eqref{eq:MP_inverse_pseudo} and to Eqn.~\eqref{eq:MP}.
\section{DACR with one type-I Segment}

This type features an additional degree of freedom, \textit{i.e.}, variable segment length.
A recent survey by Alondoli \textit{et al.} \cite{AlandoliFanLiu_Robotica_2024} shows that this robot morphology gains more relevance.
In the following, we discuss how the variable segment length is incorporated using the current representation.
Finally, we advocate to include a third variable.

\subsection{Incorporating Variable Segment Length}

First, we restate the relation between arc parameters and Clarke coordinates.
Second, we reestablish the connection between displacements $\rhovec\kthsegment{j}$ and Clarke coordinates $\rhoclarke\kthsegment{j}$.
For that, we can conclude that neither $\rhovec\kthsegment{j}$ nor $\rhoclarke\kthsegment{j}$ can recover segment length $l\kthsegment{j}$.
Afterward, two filter properties are restated to infer the segment length $l\kthsegment{j}$ from the joint length as input.

For the following, we assume a symmetric joint arrangement as stated in Eqn.~\eqref{eq:joint_location_symmetric}. 
The segment length $l\kthsegment{j}$ is encoded in the Clarke coordinates if constant curvature is assumed to establish the connection to the arc parameters \cite{GrassmannSenykBurgner-Kahrs_arXiv_2024}. 
The non-linear combination of the arc parameters is
\begin{align}
    \begin{bmatrix}
        d\kthsegment{j}l\kthsegment{j}\kappa\kthsegment{j}\cos\left(\theta\kthsegment{j}\right) \\ 
        d\kthsegment{j}l\kthsegment{j}\kappa\kthsegment{j}\sin\left(\theta\kthsegment{j}\right)
    \end{bmatrix}
    = 
    \rhoclarke\kthsegment{j}
    = 
    \boldsymbol{M}_\mathcal{P}
    \rhovec\kthsegment{j}
    ,
    \label{eq:rhoclarke_arc_parameters_constant_curvature}
\end{align}
where $\theta\kthsegment{j}$ is the bending plane angle, $\phi\kthsegment{j} = l\kthsegment{j}\kappa\kthsegment{j}$ is the bending angle, and $\kappa\kthsegment{j}$ is the curvature within in the bending plane.
Note that $\kappa\kthsegment{j}$ is constant \textit{w.r.t.} the arc length $s$.
Therefore, the constant curvature assumption \cite{JonesWalker_TRO_2006} is necessary to derive the left-hand side of Eqn.~\eqref{eq:rhoclarke_arc_parameters_constant_curvature}.
However, the right-hand side does not rely on this assumption \cite{GrassmannSenykBurgner-Kahrs_arXiv_2024, GrassmannBurgner-Kahrs_arXiv_2024b, GrassmannBurgner-Kahrs_arXiv_2025}.

The magnitude of $\rhovec\kthsegment{j}$ and $\rhoclarke\kthsegment{j}$ are related \cite{GrassmannBurgner-Kahrs_arXiv_2024a, GrassmannBurgner-Kahrs_ICRA_EA_2024}, \textit{i.e.},
\begin{align}
    \left(\rhoclarke\kthsegment{j}\right)\transpose\rhoclarke\kthsegment{j}
    &=
    \dfrac{2}{n\kthsegment{j}}\left(\rhovec\kthsegment{j}\right)\transpose\rhovec\kthsegment{j}
    .
    \nonumber
\end{align}
Given this relation, in principle, the $l\kthsegment{j}\kappa\kthsegment{j}$ can be found for a given $\rhovec\kthsegment{j}$.
However, the segment length $l\kthsegment{j}$ and the curvature $\kappa\kthsegment{j}$ are combined in a non-linear way, \textit{i.e.}, $l\kthsegment{j}\kappa\kthsegment{j}$.
Therefore, $l\kthsegment{j}$ cannot be computed without prior knowledge of $\kappa\kthsegment{j}$.
Especially for the constant curvature case, $l\kthsegment{j}$ cannot be computed in the straight configuration.
For the non-constant curvature case, if $\phi\kthsegment{j} = 0$, both $\rhoclarke\kthsegment{j}$ and $\rhovec\kthsegment{j}$ are zero vectors \cite{GrassmannBurgner-Kahrs_arXiv_2024b}.
As a consequence, neither $\rhovec\kthsegment{j}$ nor $\rhoclarke\kthsegment{j}$ can be used to compute $l\kthsegment{j}$.

This is already hinted by the fact that the displacements \eqref{eq:rho} are decoupled from the segment length $l\kthsegment{j}$.
The fact that $l\kthsegment{j}$ cannot be recovered is indicated by the properties
\begin{align}
    \!\!
    \mathbf{0}_{2 \times 1} = \MP\onevec_{n \times 1}
    \!\quad\text{and}\quad\!
    \mathbf{0}_{n \times 1} = \MPinv\MP\onevec_{n \times 1}
    ,
    \label{eq:properties}
\end{align}
where $\onevec_{n \times 1}$ has ones everywhere, whereas $\mathbf{0}_{2 \times 1}$ and $\mathbf{0}_{n \times 1}$ have zeros everywhere.
For the sake of readablity and compactness, we omit the reference to $j$ in $n\kthsegment{j}$ for the subscriptions, \textit{e.g.}, $\onevec_{n \times 1}$ instead of $\onevec_{n\kthsegment{j} \times 1}$.
Both properties \eqref{eq:properties} show that a constant value, \textit{e.g.}, $l\kthsegment{j}\onevec_{n \times 1}$, is filtered out.
However, Eqn.~\eqref{eq:properties} are useful to infer $l\kthsegment{j}$ from the joint length.

In the literature, \textit{e.g.}, \cite{WebsterJones_IJRR_2010, RaoBurgner-Kahrs_et_al_Frontiers_2021}, the joint lengths $l\kthsegment{j}\kthjoint{i} = l\kthsegment{j} - \rho\kthsegment{j}\kthjoint{i}$ are often used. 
Vectorizing $l\kthsegment{j}\kthjoint{i}$ leads to
\begin{align}
    \qv\kthsegment{j}
    =
    \begin{bmatrix} 
        l\kthsegment{j}\kthjoint{1} & l\kthsegment{j}\kthjoint{2} & \hdots & l\kthsegment{j}\kthjoint{n}
    \end{bmatrix}
    \transpose
    =
    l\kthsegment{j}\mathbf{1}_{n \times 1} - \rhovec\kthsegment{j}
    ,
	\label{eq:q}
\end{align}
where $l\kthsegment{j}\mathbf{1}_{n \times 1}$ equally adds $l\kthsegment{j}$ to all displacements.
Using the properties \eqref{eq:properties} stated above and in \cite{GrassmannBurgner-Kahrs_ICRA_EA_2024, GrassmannBurgner-Kahrs_arXiv_2025}, we can find $\rhovec\kthsegment{j} = \MPinv\MP\qv\kthsegment{j}$.
Applying it to Eqn.~\eqref{eq:q} leads to 
\begin{align}
    \left(\boldsymbol{I}_{n \times n} + \MPinv\MP\right)\qv\kthsegment{j} = l\kthsegment{j}\mathbf{1}_{n \times 1}
    \label{eq:l_ones}
\end{align}
after some algebraic manipulation. 
To extract $l\kthsegment{j}$, one might use left multiplication with $\mathbf{1}_{1 \times n}$, which leads to 
\begin{align}
    \mathbf{1}_{1 \times n}\left(\boldsymbol{I}_{n \times n} + \MPinv\MP\right)\qv\kthsegment{j} = n\kthsegment{j}l\kthsegment{j}
    \label{eq:l_from_l_ones}
    ,
    % \label{eq:}
\end{align}
or the dot product of Eqn.~\eqref{eq:l_ones}.

\subsection{Joint Representation including Length Extension}

Clarke coordinates \eqref{eq:rho_clarke} are two variables related to two arc parameters, \textit{i.e.}, $\phi\kthsegment{j}\cos\left(\theta\kthsegment{j}\right)$ and $\phi\kthsegment{j}\sin\left(\theta\kthsegment{j}\right)$, as indicated by Eqn.~\eqref{eq:rhoclarke_arc_parameters_constant_curvature}.
Therefore, it is unsurprising that the Clarke coordinates, \textit{i.e.}, $\rhoreal\kthsegment{j}$ and $\rhoim\kthsegment{j}$, cannot account for a third variable, \textit{e.g.}, segment length $l\kthsegment{j}$.

While joint length \eqref{eq:q} can be used to find the segment length $l\kthsegment{j}$, it simplifies to displacement \eqref{eq:rho} for type-0 segments.
This is due to Eqn.~\eqref{eq:properties}.
Indeed, all proposed improved state parameterizations \cite{DellaSantinaBicchiRus_RAL_2020, AllenAlbert_et_al_RoboSoft_2020, CaoXie_et_al_JMR_2022} derived for $\qv$ can be derived using $\rhovec$, and are Clarke coordinates \cite{GrassmannBurgner-Kahrs_arXiv_2025}.
To take advantage of the displacement representation, we can write 
\begin{align}
    \begin{bmatrix}
        \rhoclarke\kthsegment{j}\\[0.5em]
        \beta\kthsegment{j}
    \end{bmatrix}
    =
    \begin{bmatrix}
        \MP & \mathbf{0}_{n \times 1}\\[0.5em]
        \mathbf{0}_{1 \times n} & 1
    \end{bmatrix}
    \begin{bmatrix}
        \rhovec\kthsegment{j}\\[0.5em]
        \beta\kthsegment{j}
    \end{bmatrix}
    ,
    \label{eq:forward_type-1}
\end{align}
as transformation between both extended parameterizations.
Both are extended by $\beta\kthsegment{j}$, a joint value of $j\textsuperscript{th}$ segment length. 
When using $\qv\kthsegment{j}$ instead of $\rhovec\kthsegment{j}$, one can substitute $\rhovec\kthsegment{j}$ for $-\qv\kthsegment{j}$ in Eqn.~\eqref{eq:forward_type-1}. 
Alternatively, we can find 
\begin{align}
    \begin{bmatrix}
        \rhoclarke\kthsegment{j}\\[0.5em]
        \beta\kthsegment{j}
    \end{bmatrix}
    =
    \begin{bmatrix}
        -\MP \\[0.5em]
        \dfrac{1}{n\kthsegment{j}}\mathbf{1}_{1 \times n}\left(\boldsymbol{I}_{n \times n} + \MPinv\MP\right)
    \end{bmatrix}
    \qv\kthsegment{j}
    ,
    \label{eq:forward_type-1_q}
\end{align}
where $\beta\kthsegment{j} = l\kthsegment{j}$ and Eqn.~\eqref{eq:l_from_l_ones} are used.
Note that the used convention in Eqn.~\eqref{eq:q} causes a negative $\MP$ in Eqn.~\eqref{eq:forward_type-1_q}.

The right selection between Eqn.~\eqref{eq:forward_type-1} and Eqn.~\eqref{eq:forward_type-1_q} is informed by the given hardware. 
For example, Eqn.~\eqref{eq:forward_type-1} is useful for tendon-driven continuum robot, \textit{e.g.}, \cite{NguyenBurgner-Kahrs_IROS_2015}, as the variable length is actuated by an additional actuator. 
In contrast, a pneumatic actuated continuum soft robot, \textit{e.g.}, \cite{DellaSantinaBicchiRus_RAL_2020},  is better described by Eqn.~\eqref{eq:forward_type-1_q} as the variable length is an intrinsic feature of pneumatic actuators.

The inverse of Eqn.~\eqref{eq:forward_type-1} is straightforward as it just involves $\MPinv$, \textit{cf.} Eqn.~\eqref{eq:inverse} and Eqn.~\eqref{eq:forward_type-1}.
To find the inverse of Eqn.~\eqref{eq:forward_type-1_q}, we can vectorize Eqn.~\eqref{eq:q} instead of finding the pseudoinverse of the matrix used in Eqn.~\eqref{eq:forward_type-1_q}.
This leads directly to 
\begin{align}
    \qv\kthsegment{j}
    =
    \begin{bmatrix}
        -\MPinv & \ \mathbf{1}_{n \times 1}
    \end{bmatrix}
    \begin{bmatrix}
        \rhoclarke\kthsegment{j}\\[0.5em]
        \beta\kthsegment{j}
    \end{bmatrix}
    \label{eq:inverse_typ-1_q}
    ,
\end{align}
where $\beta\kthsegment{j}$ is $l\kthsegment{j}$.
Note that $\left[-\MPinv; \mathbf{1}_{n \times 1}\right] \in \mathbb{R}^{n \times 3}$.
\section{Other types of DACR}

In this section, we briefly dive into further extension as mentioned in Sec.~\ref{sec:DACR}.
In particular, DACR with one type-II segment and DACR with one type-III segment as well as DACR with multiple segments.
We acknowledge that those abstractions are increasingly dependent on the hardware of the continuum robot.
Therefore, a deep dive and over-generalization might not be appropriate. 
Nevertheless, some intuition and formulae are provided to guide future work. 
Furthermore, for the sake of brevity, our exploration is limited to the symmetric arrangement \eqref{eq:joint_location_symmetric}.

\subsection{Incorporating Twisting Segment}

\begin{figure}
    \centering
    \includegraphics[width=\columnwidth]{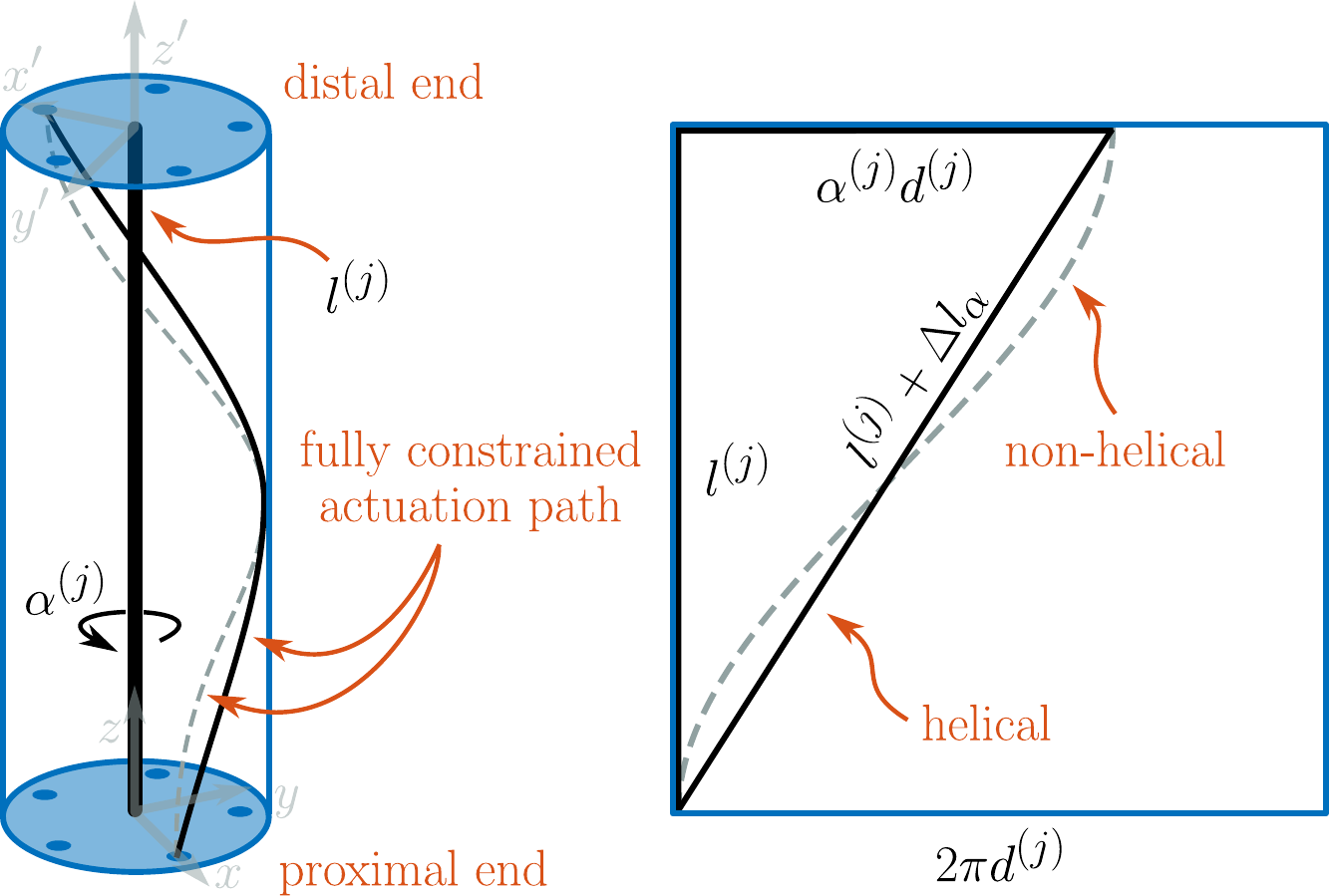}
    \caption{
        Helical fully constrained actuation path.
        This actuation path lies in a plane that can be rolled out, revealing a linear function and a Pythagorean equation.
    }
    \label{fig:helical-actuation-path}
\end{figure}

Assuming a fully constrained actuation path, one can show that a rotational joint $\alpha\kthsegment{j}$ can create a helical actuation path \cite{GrassmannBurgner-Kahrs_et_al_Frontiers_2022}.
Figure~\ref{fig:helical-actuation-path} illustrates a helical fully constrained actuation path.
Using straightforward geometry depicted in Fig.~\ref{fig:helical-actuation-path}, a Pythagorean equation relates $\alpha\kthsegment{j}$ to an increase of displacement $\Delta l_\alpha$.
After rearranging, one can state
\begin{align}
    \Delta l_\alpha = \sqrt{\left(\alpha\kthsegment{j} d\kthsegment{j}\right)^2 + \left(l\kthsegment{j}\right)^2} - l\kthsegment{j}
    \nonumber
    .
\end{align}
This offset induced by $\alpha\kthsegment{j}$ is a constant equally applied to all displacement-actuated joints in $j\textsuperscript{th}$.
This is similar to $l\kthsegment{j}$ for $\qv\kthsegment{j}$ in Eqn.~\eqref{eq:q}.
Therefore, $\Delta l_\alpha$ vanishes when applying the Clarke transform due to Eqn.~\eqref{eq:properties}.

Relaxing the assumption on the helical actuation path, we can say that any function, \textit{i.e.}, $\Delta l_\alpha = f\left(\alpha\kthsegment{j}, s\right)$, will map to zero if $\Delta l_\alpha$ is an additive constant applied to Eqn.~\eqref{eq:q}.
Recap the properties \eqref{eq:properties}.
Figure~\ref{fig:helical-actuation-path} shows a non-helical example as well.

\subsection{DACR with one Type-II or Type-III Segment}

For DACR with one type-III segment, we can state two transformations.
As $l\kthsegment{j}$ is not affected by $\alpha\kthsegment{j}$ and $\Delta l_\alpha$ is an additive offset, we can simply extend both representations and the matrix.
To include $\alpha$ in Eqn.~\eqref{eq:forward_type-1}, this leads to
\begin{align}
    \begin{bmatrix}
        \rhoclarke\kthsegment{j}\\[0.25em]
        \beta\kthsegment{j}\\
        \alpha\kthsegment{j}
    \end{bmatrix}
    =
    \begin{bmatrix}
        -\MP & \mathbf{0}_{n \times 2}\\[0.5em]
        \mathbf{0}_{2 \times n} & \Imat_{2 \times 2}
    \end{bmatrix}
    \begin{bmatrix} 
        \qv\kthsegment{j}\\[0.25em]
        \beta\kthsegment{j}\\
        \alpha\kthsegment{j}
    \end{bmatrix}
    ,
    \label{eq:forward_type-3}
\end{align}
where $\qv\kthsegment{j} = \left(l\kthsegment{j} + \Delta l_\alpha\right)\mathbf{1}_{n \times 1} - \rhovec\kthsegment{j}$ is an extension of Eqn.~\eqref{eq:q}.
A similar padding is used for Eqn.~\eqref{eq:forward_type-1_q}.
This leads to
\begin{align}
    \!\!
    \begin{bmatrix}
        \rhoclarke\kthsegment{j}\\[0.25em]
        \beta\kthsegment{j}\\
        \alpha\kthsegment{j}
    \end{bmatrix}
    \!\!
    =
    \!\!
    \begin{bmatrix}
        -\MP  & \mathbf{0}_{n \times 1}\\[0.5em]
        \dfrac{1}{n\kthsegment{j}}\mathbf{1}_{1 \times n}\left(\boldsymbol{I}_{n \times n} + \MPinv\MP\right)  & 0\\[0.5em]
        \mathbf{0}_{1 \times n} & 1
    \end{bmatrix}
    \!\!
    \begin{bmatrix}
        \qv\kthsegment{j}\\[0.25em]
        \alpha\kthsegment{j}
    \end{bmatrix}
    \!\!
    ,
    \nonumber
\end{align}
where $\qv\kthsegment{j}$ is compensated for $\Delta l_\alpha$ in order to achieve Eqn.~\eqref{eq:q}.
Otherwise, Eqn.~\eqref{eq:l_ones} does not hold and $\beta\kthsegment{j}$ depends on $\alpha\kthsegment{j}$.
Further intuition on the compensation is given in Sec.~\ref{sec:multiple_segments}.

Regarding DACR with one type-II segment, we can simplify Eqn.~\eqref{eq:forward_type-3} accordingly.
This is equivalent to substitute $\beta\kthsegment{j}$ with $\alpha\kthsegment{j}$ in Eqn.~\eqref{eq:forward_type-1}.

Their inverse mapping is straightforward, too. 
One can compare Eqn.~\eqref{eq:forward_type-3} with Eqn.~\eqref{eq:forward_type-1}, which is resemble Eqn.~\eqref{eq:forward}.
For a DACR with one type-III segment, the resulting inverse for the above mapping resembles Eqn.~\eqref{eq:inverse_typ-1_q}.

\subsection{Multiple Segments}
\label{sec:multiple_segments}

Consider a DACR with two type-0 segments.
Each segment has symmetric joint arrangements, see Eqn.~\eqref{eq:joint_location_symmetric}.
We distinguish two prominent cases.
First, all segments are interdependent actuated, \textit{i.e.}, actuators of the approximal segment cannot influence the distal segment and \textit{vice versa}.
This is the case for pneumatic actuated joints.
Second, actuators of the distal segment cause the bending of the proximal segment.
This is the case for tendon-driven continuum robots, where the tendons of the distal segment are routed through the proximal segment.

For the first case, \textit{i.e.}, independent segments, Eqn.~\eqref{eq:forward} can be applied independently for $\rhovec\kthsegment{1}$ and $\rhovec\kthsegment{2}$ or for $-\qv\kthsegment{1}$ and $-\qv\kthsegment{2}$.
A diagonal matrix with block matrix $\MP$ as diagonal elements can be used for a compact transformation.

For the second case, \textit{i.e.}, interdependent segments, the difference between the actuation and joint space needs to be clear.
This also affects the definition of $f_\text{dyn}$ and $f_\text{dyn}^{-1}$, see Fig.~\ref{fig:abstraction_overview}.
One might consider two approaches to solve the boundary between the abstractions used for the actuation and joint space.
They are admittedly very similar.
The unifying idea is reformulating the joint length
\begin{align}
    \qv\kthsegment{1}
    &=
    l\kthsegment{1}\mathbf{1}_{n \times 1} - \rhovec\kthsegment{1}
    \qquad\qquad\text{and}
    \label{eq:q1}
    \\
    \qv\kthsegment{2}
    &=
    l\kthsegment{2}\mathbf{1}_{n \times 1} - \rhovec\kthsegment{2} + \qv\kthsegment{1}
    \label{eq:q2}
\end{align}
to the desired formulation in Eqn.~\eqref{eq:q}.
This can be done after $f_\text{dyn}$ in the joint space or as part of $f_\text{dyn}$ in the actuation space representing the two general approaches.
In the following, we only consider the former due to there similarities.

Combining Eqn.~\eqref{eq:q1} and Eqn.~\eqref{eq:q2} in vectorized form leads to
\begin{align}
    \begin{bmatrix}
        \rhoclarke\kthsegment{1}\\[0.5em]
        \rhoclarke\kthsegment{2}
    \end{bmatrix}
    &=
    \begin{bmatrix}
        \MP & \mathbf{0}_{2 \times n}\\[0.5em]
        \mathbf{0}_{2 \times n} & \MP
    \end{bmatrix}
    \begin{bmatrix}
        \rhovec\kthsegment{1}\\[0.5em]
        \rhovec\kthsegment{2}
    \end{bmatrix}
    \label{eq:forward_stack}
\end{align}
after applying \eqref{eq:forward}.
The interdependency is apparent in
\begin{align}
    \begin{bmatrix}
        \rhoclarke\kthsegment{1}\\[0.5em]
        \rhoclarke\kthsegment{2}
    \end{bmatrix}
    &=
    \begin{bmatrix}
        -\MP & \mathbf{0}_{2 \times n}\\[0.5em]
        \MP & -\MP
    \end{bmatrix}
    \begin{bmatrix}
        \qv\kthsegment{1}\\[0.5em]
        \qv\kthsegment{2}
    \end{bmatrix}
    \label{eq:forward_stack_q}
    ,
\end{align}
when using joint length $\qv\kthsegment{j}$.
Due to properties \eqref{eq:properties}, the constant terms in Eqn.~\eqref{eq:q1} and Eqn.~\eqref{eq:q2} are omitted in Eqn.~\eqref{eq:forward_stack_q}.

% Discussion
\section{Discussion}

We revisit and extend the abstraction \textit{displacement-actuated continuum robot}, which is distinct from the commonly used constant curvature abstraction in continuum robotics.
The constant curvature assumption \cite{JonesWalker_TRO_2006} is an abstraction of the arc space, while DACR is an abstraction of the joint space, see Fig.~\ref{fig:abstraction_overview}.

Our extended DACR abstration can account for segments of variable length and segments that can twist.
The former has gained more relevance recently \cite{AlandoliFanLiu_Robotica_2024}, whereas the latter is underutilized as an active degree of freedom \cite{GrassmannBurgner-Kahrs_et_al_Frontiers_2022}.
Furthermore, our extension includes their combination and multiple segments.
While a deep dive relies on the physical realization of the continuum robot, we hope to provide intuition and insight to start with.
For example, the pattern in Eqn.~\eqref{eq:forward_stack} and in Eqn.~\eqref{eq:forward_stack_q} should be evident on how $m$ interdependent segments scale.
Furthermore, the reader can extend the forward mapping to inverse mapping and arbitrary joint locations \eqref{eq:joint_location_non-symmetric} with ease.
This includes different numbers of joints for each segment.

The computation of the length $l\kthsegment{j}$ via Eqn.~\eqref{eq:l_ones} from $\qv\kthsegment{j}$ for Eqn.~\eqref{eq:forward_type-1_q} and Eqn.~\eqref{eq:forward_type-3} can be used to reduce the number of sensors as a dedicated sensor for $l\kthsegment{j}$ or $\beta\kthsegment{j}$ is not needed.
It can also detect compression or decompression, \textit{i.e.}, a divergence from a desired $l\kthsegment{j}$.
More importantly, $l\kthsegment{j}$ and $\beta\kthsegment{j}$ are linearly correlated to $\qv\kthsegment{j}$, which is, in general, an undesirable property for an input. 
Therefore, approaches utilizing, for example, a controller or a neural network, might suffer from a loss in performance.

We note that we do not solve the kinematics of the extensions.
Our presentation is limited to possibilities within the abstraction and the effect when applying the Clarke transform.
Constant offsets added to the displacements $\rhovec\kthsegment{j}$ are filtered out due to Eqn.~\eqref{eq:properties}.
For example, constant offsets induced by variable length or helical actuation path are mapped to zero.
We also present a way to recover the length from joint length $\qv\kthsegment{j}$ without using the dot product of $\qv\kthsegment{j}$.
This leads to a linear transformation \eqref{eq:forward_type-1_q}, providing possibilities to simplify mathematical formulations.
Future work will investigate the kinematics based on our extensions of the DACR abstraction.
We hypothesize that their forward and inverse kinematics can be formulated as closed-form solutions as the one shown by Grassmann \textit{et al.} \cite{GrassmannSenykBurgner-Kahrs_arXiv_2024} for a DACR with one type-0 segment.

Both assumptions for the investigated abstraction stated in Sec.~\ref{sec:DACR} lead to a lower bound for the geometric smoothness, \textit{i.e.}, the backbone is at least $\mathcal{G}^1$ smooth.
The smoothness, the application of parallel curves, and the sketch to derive the spatial case from the planar case, which also implies a derivation of Eqn.~\eqref{eq:sum_rho} from Eqn.~\eqref{eq:sum_rho_planar_case}, add to the geometric insights by Simaan \textit{et al.} \cite{SimaanTaylorFlint_ICRA_2004}, Burgner-Kahrs \textit{et al.} \cite{Burgner-KahrsRuckerChoset_TRO_2015}, and Grassmann \& Burgner-Kahrs \cite{GrassmannBurgner-Kahrs_arXiv_2024b}.
Besides the insights gained by the use of parallel curves, parallel curves provide an analytic and geometric representation to extend spatial curves to area and volume.
Note that parallel curves are not necessarily the same type as the original curve \cite{Pham_CAD_1992}, \textit{i.e.}, the curve representing the backbone.
Providing a rich source for future directions, parallel curves and their extensions would allow for effective and efficient geometric approaches for collision detection, obstacle avoidance, contact-navigated path planning \cite{RaoSalzmanBurgner-Kahrs_RAL_2024}, and motion planning based on kinematics.

Furthermore, the definition of the abstraction reveals implicit assumptions that might be hidden from the researcher and practitioner.
Note that the convention in Eqn.~\eqref{eq:q}, \textit{i.e.}, the direction of positive values, and the boundaries between actuation and joint space, as discussed in Sec.~\ref{sec:multiple_segments}, are important considerations, too.
All assumptions can also provide a source for further future directions.
For example, the tapered or cone-like continuum structures could be considered by changing constant $d\kthjoint{i}\kthsegment{j}$ to variable $d\kthjoint{i}\kthsegment{j}\!\left(s\right)$, where arc length $s$ is a parameter indicating the position on the neutral axis of the continuum structure.
However, this would require extending the Clarke transform \cite{GrassmannSenykBurgner-Kahrs_arXiv_2024} accordingly.
Another example is the limitation to kinematic parameters that assume the absence of general forces as they would require the consideration of dynamic parameters such as mass.
Therefore, deriving an equivalent abstraction that is useful for dynamics is a promising future direction.

Finally, we highlight that the improved state parameterizations proposed by Della \textit{et al.} \cite{DellaSantinaBicchiRus_RAL_2020}, Allen \textit{et al.} \cite{AllenAlbert_et_al_RoboSoft_2020}, and Dian \textit{et al.} \cite{DianGuo_et_al_Access_2022} are specific instances of the Clarke coordinates \cite{GrassmannBurgner-Kahrs_arXiv_2025}.
Furthermore, Clarke coordinates \cite{GrassmannSenykBurgner-Kahrs_arXiv_2024} emerge directly from displacement representation \eqref{eq:rho} and the abstraction of the joint space, see Fig.~\ref{fig:abstraction_overview}.
Since Clarke coordinates are a representation of the joint space, the improved state parameterizations are representations of the joint space and not necessarily a representation of the arc space.
Note that substituting $\theta\kthsegment{j}$ with $l\kthsegment{j}\kappa\kthsegment{j}$ implies the use of constant curvature assumption.
Therefore, improved state parameterizations are not limited to constant curvature assumption \cite{WebsterJones_IJRR_2010}, making them more general than previously assumed and presented.

% Conclusion
\section{Conclusion}

We investigate an abstraction called displacement-actuated continuum robots and introduce their extensions to consider length extension and twist.
Furthermore, the corresponding Clarke transform is provided.
Those abstractions provide the possibility to generalize and unify approaches.

%%%%%%%%%%%%%%%%%%%%%%%%%%%%%%%%%%%%%%%%%%%%%%%%%%%%%%%%%%%%%%%%%%%%%%
\begin{acknowledgment}
We acknowledge the support of the Natural Sciences and Engineering Research Council of Canada (NSERC), [RGPIN-2019-04846].
\end{acknowledgment}

%%%%%%%%%%%%%%%%%%%%%%%%%%%%%%%%%%%%%%%%%%%%%%%%%%%%%%%%%%%%%%%%%%%%%%
% The bibliography is stored in an external database file
% in the BibTeX format (file_name.bib).  The bibliography is
% created by the following command and it will appear in this
% position in the document. You may, of course, create your
% own bibliography by using thebibliography environment as in
%
% \begin{thebibliography}{12}
% ...
% \bibitem{itemreference} D. E. Knudsen.
% {\em 1966 World Bnus Almanac.}
% {Permafrost Press, Novosibirsk.}
% ...
% \end{thebibliography}

% Here's where you specify the bibliography style file.
% The full file name for the bibliography style file 
% used for an ASME paper is asmems4.bst.
\bibliographystyle{asmems4}

% Here's where you specify the bibliography database file.
% The full file name of the bibliography database for this
% article is asme2e.bib. The name for your database is up
% to you.
\bibliography{literature}

% %%%%%%%%%%%%%%%%%%%%%%%%%%%%%%%%%%%%%%%%%%%%%%%%%%%%%%%%%%%%%%%%%%%%%
% \appendix       %%% starting appendix
% \section*{Appendix A: Head of First Appendix}
% Avoid Appendices if possible.

% %%%%%%%%%%%%%%%%%%%%%%%%%%%%%%%%%%%%%%%%%%%%%%%%%%%%%%%%%%%%%%%%%%%%%%
% \section*{Appendix B: Head of Second Appendix}
% \subsection*{Subsection head in appendix}
% The equation counter is not reset in an appendix and the numbers will
% follow one continual sequence from the beginning of the article to the very end as shown in the following example.
% \begin{equation}
% a = b + c.
% \end{equation}

\end{document}